\documentclass[conference]{IEEEtran}

\usepackage{graphicx}
\usepackage{amsmath, amssymb}
\usepackage{booktabs}
\usepackage{multirow}
\usepackage{longtable}
\usepackage{soul}
\usepackage{color}
\usepackage[table]{xcolor}
\usepackage{colortbl}
\usepackage{xcolor}
\usepackage[hidelinks]{hyperref}
\definecolor{bestrow}{RGB}{213, 240, 213}  
\usepackage[hidelinks]{hyperref}

\title{SwiftCTS: Fast Cross-Design Prediction and Pareto Optimization of Clock Tree Metrics via Few-Shot Calibration}

\author{
\IEEEauthorblockN{Barsat Khadka}
\IEEEauthorblockA{The University of Southern Mississippi \\
Hattiesburg, MS, USA \\
Barsat.Khadka@usm.edu}
\and
\IEEEauthorblockN{Kawsher Ahmed Roxy}
\IEEEauthorblockA{Intel Corporation\\
Hillsboro, OR, USA \\
kawsher.roxy@intel.com}
\and
\IEEEauthorblockN{Md Rubel Ahmed}
\IEEEauthorblockA{Louisiana Tech University \\
Ruston, LA, USA \\
mahamed@latech.edu}
}

\begin{document}

\maketitle

\begin{abstract}
Clock Tree Synthesis (CTS) is a computationally expensive stage in the physical design flow, requiring iterative EDA tool invocations to navigate a vast configuration space for optimal power, wirelength, and timing skew. Existing machine learning approaches require computationally expensive retraining or fine-tuning cycles to adapt to unseen macro architectures and are architecturally mismatched to the millions of evaluations demanded by exhaustive combinatorial search. We present SwiftCTS, a physics-informed surrogate framework that addresses both limitations simultaneously. By coupling lightweight, physics-grounded statistical features with gradient-boosted ensembles, SwiftCTS trains in under five seconds on a CPU and delivers sub-millisecond inference without GPU support. To handle out-of-distribution (OOD) designs without retraining or fine-tuning, we introduce a K-shot multiplicative calibration mechanism that anchors predictions to just one or two physical reference runs, reducing power prediction error from 24.5\% to 3.3\% and wirelength error from 56.6\% to under 1\% on unseen macros. Integrating this engine with an evolutionary optimizer, SwiftCTS evaluates 100,000 CTS configurations in under ten seconds, yielding Pareto-optimal frontiers that are physically validated within the OpenROAD flow. Closed-loop validation confirms prediction errors below 0.5\% for power and wirelength, and timing skew predictions within five picoseconds on an OOD benchmark, consistently outperforming default tool heuristics across all target metrics. Code publicly available at: \href{https://anonymous.4open.science/r/SwiftCTS-7E6E}{https://github.com/BarsatKhadka/SwiftCTS}
\end{abstract}

\section{Introduction}

Clock Tree Synthesis remains a computationally expensive stage in the modern physical design flow. Because the clock distribution network significantly influences overall chip power, timing skew, and routing congestion, engineers must carefully navigate a vast configuration space via repeatedly invoking the EDA tool on a fixed placement to evaluate different configurations and  achieve optimal Quality of Results (QoR). This exhaustive trial-and-error workflow imposes a massive computational bottleneck and severely prolongs the physical design cycle. To automate this task, various methods have been proposed, ranging from rule-based data mining \cite{kahng}, high-dimensional metamodeling \cite{kahng2, kahng3}, neural networks \cite{kwon2018}, to spatial and generative architectures \cite{nagaria2020} \cite{gan-cts}. Early heuristic and metamodeling approaches were fast as they relied on aggregate or hand-engineered features to approximate CTS behavior \cite{kahng, kahng2, kahng3}. As a result, their predictive accuracy degraded significantly on unseen designs. Modern deep learning and generative architectures \cite{nagaria2020, gan-cts} \cite{layout2020} resolve this accuracy deficit on isolated benchmarks, but they lack the computational flexibility required for cross-design environments as physical design frequently involves updated constraints and unseen OOD macro architectures. It is a well-documented phenomenon in ML for EDA that deep neural networks suffer with poor generalization when applied to new designs \cite{nsfReport} \cite{zhengDAC} \cite{ghoseICCAD}. Recovering accuracy requires generating massive new datasets and executing computationally expensive retraining cycles. Furthermore, their inference latency is too high to support the millions of evaluations required for exhaustive design space exploration. As a consequence, these frameworks are restricted to making few predictions based on prior learning, rather than actively evaluating the combinatorial search space to find the true global optimum \cite{gan-cts}.

\begin{figure*}[t]
    \centering
    \includegraphics[width=0.85\textwidth]{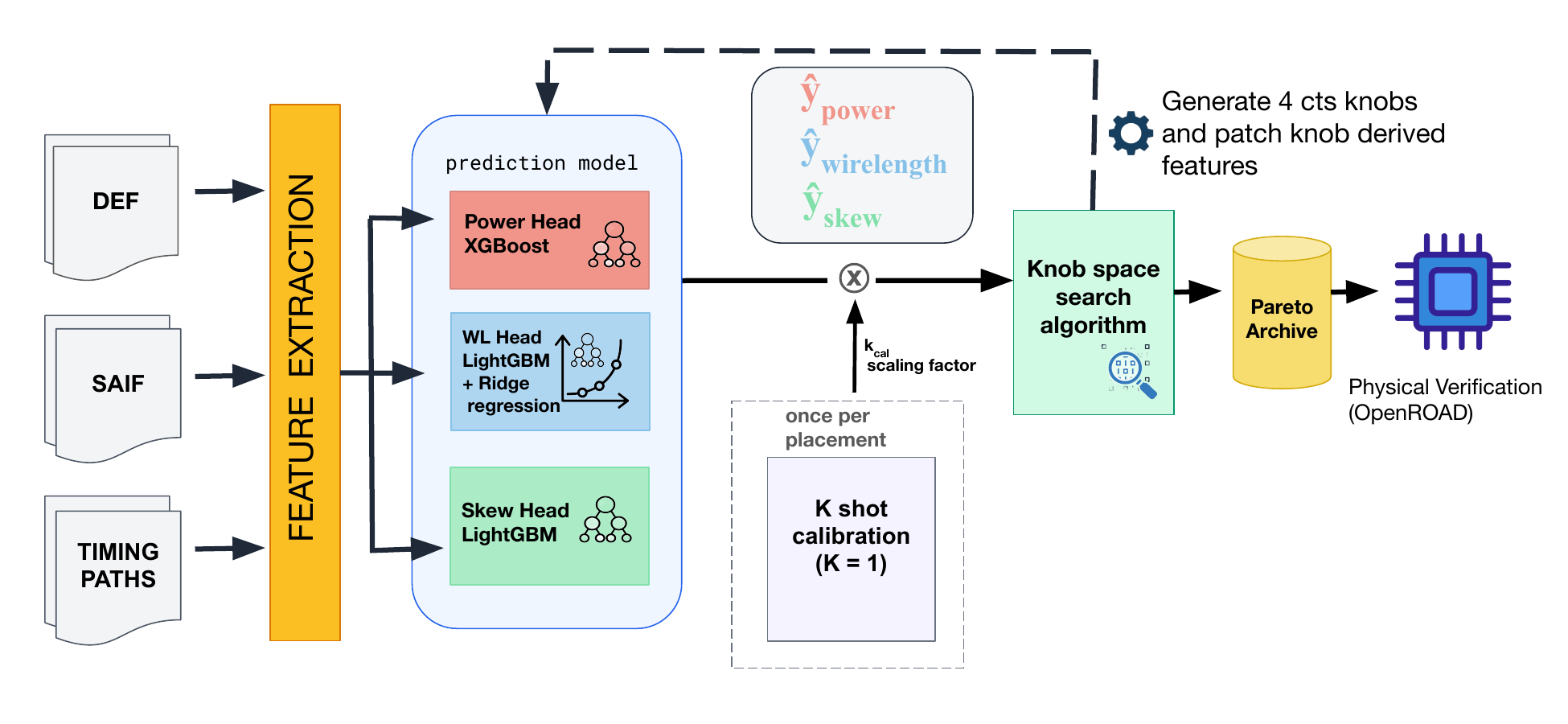} 
    \caption{High-level overview of the SwiftCTS framework. The pipeline begins with feature extraction from placement data, followed by a lightweight surrogate model for rapid CTS metric prediction. A K-shot calibration module aligns predictions to new designs, and an NSGA-II optimizer explores the configuration space to generate Pareto-optimal solutions.}
    \label{fig:architecture}
\end{figure*}

This fundamental limitation raises a critical question: \textit{Is it possible to design a surrogate framework that adapts near-instantaneously to new designs and possesses the inference speed to evaluate millions of configurations in minutes?} To answer this, we introduce SwiftCTS, a high-speed surrogate framework designed to accurately predict clock tree metrics and drive agile Design Space Exploration (DSE). The primary contributions of this paper are as follows:
\begin{itemize}
    \item[$\bullet$] \textbf{Agile Surrogate Architecture:} We propose an ultra-fast surrogate model that pivots away from heavy deep neural networks. By utilizing gradient-boosted decision trees, SwiftCTS achieves sub-millisecond inference and allows for full tree ensembles to be retrained from scratch in seconds, bypassing the need for extensive GPU time.
    \item[$\bullet$] \textbf{K-Shot Offset Calibration:} We introduce a novel calibration technique to handle unseen macro architectures without full retraining. By evaluating only one or two baseline physical runs of a new placement, the surrogate mathematically aligns its predictions to the new domain while preserving the core model weights.
    \item[$\bullet$] \textbf{Search-Based Pareto Optimization:} We integrate our rapid prediction engine with an multi-objective evolutionary optimizer. By utilizing the surrogate to instantly predict CTS outcomes across millions of heuristic configurations, the framework actively sweeps the design space in minutes to yield strictly dominant Pareto frontiers.
\end{itemize}
The rest of this paper is organized as follows: Section II discusses related work. Section III details the SwiftCTS framework. Section IV and Section V present the experimental setup and results, respectively. Section VI provides a discussion, and finally we conclude our paper in Section VII.

\section{Related Work}

To automate design space exploration, various machine learning techniques have been proposed. These approaches can be broadly categorized by their architectural priorities:

\textit{1) Aggregate-Feature Analytical Modeling:} Early works \cite{kahng} utilized rule-based data mining to estimate clock skew and insertion delay. Subsequent research \cite{kahng2, kahng3} employs high-dimensional metamodeling to predict more complex objectives, such as clock power and wirelength, while accounting for non-uniform sink distributions and varying floorplan aspect ratios. While efficient, they relied on aggregate or hand-engineered features such as gate count, routing layers, whitespace, and floorplan aspect ratios to approximate CTS behavior. As a result, their predictive accuracy degraded significantly on unseen designs, with reported correlation dropping from near-perfect ($1.0$) to as low as $0.48$ and relative errors exceeding $20\%$ when representative training instances were unavailable.

\textit{2) Deep Learning Architectures:} To overcome the limitations of analytical models, recent literature shifted toward deep learning architectures. Authors of \cite{nagaria2020} employ Convolutional Neural Networks (CNNs) to estimate CTS parameters directly from pre-CTS placement images. Taking this visual approach a step further, Lu et al. \cite{gan-cts} introduce a generative adversarial framework (GAN-CTS). Specifically, their framework relies on the EDA tool to perform placement and trial routing, after which it exports high-resolution visual maps of the flip-flop distribution, clock net distribution, and routing congestion,  which are then processed through a 50-layer convolutional network to extract spatial features. By coupling a conditional GAN with a pretrained regression supervisor on top of these features, their model actively recommends CTS input parameters to optimize power and wirelength. More recently, efforts to mitigate the severe generalization drop \cite{zhang2024} \cite{zhao2025} have explored complex transfer learning, such as disentangling node-dependent and design-dependent features to align data across different technology nodes. Despite high accuracy, a fundamental downside of all these architectures is their massive data dependency and reliance on heavy GPU acceleration.

\textit{3) Reinforcement Learning and Gen/Agentic AI:} A parallel line of research treats physical design optimization as a sequential decision-making problem. For instance, deep reinforcement learning (RL) frameworks \cite{thomas2025} \cite{agnesina} have been used to tune placement parameters by utilizing Graph Neural Networks and Transformers for multi-stage tuning. To bypass the heavy data requirements of RL, researchers have also applied Bayesian Optimization (BO) to navigate these complex parameter spaces \cite{geng2022, zheng2023}. Finally, LLM based agents \cite{chateda} \cite{spec2rtl} have automated the broader workflow, orchestrating tasks from RTL to GDSII. Despite all this, the fundamental limitation of RL, BO, and AI frameworks is that they keep the actual EDA tool inside their optimization loop. Every time the agent needs to evaluate a reward or test a new parameter, it must execute a physical tool. Because of this, these methods are restricted to finding a single good trajectory and simply cannot afford to exhaustively sweep millions of configurations.

\section{The SwiftCTS Framework}
In this section, we present a methodical overview of the proposed SwiftCTS architecture. The framework takes DEF, SAIF and Timing-path reports as inputs and outputs the Pareto-optimal CTS configurations alongside their predicted quality metrics. The core components of SwiftCTS include a shared feature extractor, independent prediction heads, a K-shot calibration module to resolve out-of-distribution (OOD) domain shift, and a multi-objective evolutionary optimizer. For any given placement, the feature engine parses the input reports exactly once to capture die geometry, cell mix, drive strength, and switching activity, constructing an exhaustive 110-dimensional physical feature space (detailed in Table \ref{tab:all_features_and_nom}). The mathematical formulations in Table \ref{tab:all_features_and_nom} are specifically engineered for scale-invariance to handle severe domain shift. Logarithmic scaling is utilized to compress features where the high dynamic range of circuit sizes is critical, while dimensionless ratios enable the surrogate to learn universal density patterns that generalize regardless of absolute design magnitude. Explicit interaction terms between layout topology and CTS tuning knobs ground the model in the underlying physics of the design. Finally, we utilize three independent heads for prediction, each with its own tailored feature space (detailed Table \ref{tab:all_features_and_nom}) and learner architecture designed to capture the specific physics of its target objective.

\subsection{Prediction Heads Setup}
Unlike shared-trunk architectures commonly used in deep learning \cite{gan-cts}, SwiftCTS employs three independent prediction heads. This design choice is motivated by the nature of gradient-boosted decision trees, which optimize splits based on gradients and variance statistics of a single target at a time. Extending a shared-tree structure to multiple objectives would require aggregating these statistics across targets, which can bias split selection toward objectives with larger variance or scale, potentially reducing sensitivity to other metrics. To avoid such interference, we train separate tree ensembles for each target while using a common feature representation. This ensures that each model can learn task-specific partitioning of the feature space without compromise.
\subsubsection{Power Prediction Head}
The power prediction head employs an XGBoost ensemble \cite{xgboost} operating on a 20-dimensional feature space. Within our evaluated benchmark suite, clock tree power varies by 8.9$\times$ across different design families, yet only by 10--30\% within a single design across different CTS knob configurations. Predicting power of an entirely unseen design family is inherently difficult because its absolute scale is completely unknown, and its sensitivity to different CTS knobs varies fundamentally by design. The power head solves this with three mechanisms:  a ratio-regression target, a physics based normalizer and a carefully constructed feature space. First, the target is transformed into a scale-free log-ratio:
\begin{equation}
y_{\text{power}} = \log\left(\frac{\text{power}}{\text{pw\_norm}}\right)
\end{equation}
where the normalizer 
$\text{pw\_norm} = n_\text{ff} \times f_\text{GHz} \times \text{avg\_ds}$ 
serves as a first-principles baseline estimate, inspired directly from the dynamic clock power equation 
\[
P = \alpha \times C_\text{total} \times V^2 \times f.
\]
By predicting the log-ratio, the ensemble is strictly tasked with regressing  the specific $10\text{--}30\%$ power variation within a design rather than memorizing absolute per-design baselines. 

\subsubsection{Wirelength Prediction Head}
 Wirelength can be analytically lower-bounded using established VLSI routing theory \cite{steinerTree}. We exploit this property to build a scale-invariant target that decouples macroscopic chip dimensions from the microscopic routing decisions made by the CTS tool.
Drawing from BHH Theorem and Euclidean geometry, the minimum Steiner tree for $N$ sinks distributed over an area $A$ scales as $\sqrt{N \times A}$. We normalize it as:
\begin{equation}
\mathrm{wl}_{\mathrm{norm}} \propto \sqrt{n_{\mathrm{ff}} \cdot A}
\end{equation}
which approximates the theoretical minimum wirelength independent of actual flip-flop coordinates or CTS configuration.  Then we define the wirelength target as $y_{\mathrm{wl}} = \log(\mathrm{WL}_{\mathrm{actual}} / \mathrm{wl}_{\mathrm{norm}})$, and the absolute wirelength is recovered via $\mathrm{WL}_{\mathrm{pred}} = \exp(\hat{y}_{\mathrm{wl}})\,\mathrm{wl}_{\mathrm{norm}}$. By predicting the relative routing penalty, we ensure that the model can seamlessly generalize across designs of vastly different scales.

The learning architecture for this head blends LightGBM \cite{lightgbm} and Ridge Regression \cite{ridgeReg} (using a 30/70 split) operating on a 75-dimensional feature space (detailed in Table \ref{tab:all_features_and_nom}). Clock wirelength exhibits a strong continuous correlation with the physical spread and congestion of the layout. However, standard decision trees like LightGBM struggle to capture continuous linear trends \cite{struggle}. On our largest training design (ETHMAC), a pure tree model flatlines and severely underpredicts the routing penalty. The Ridge Regression component is explicitly introduced to solve this linear size-scaling, while LightGBM refines the non-linear, localized heuristics. However, using linear models like Ridge Regression during inference on a completely unseen design that far exceeds the bounds of the training data can lead to severe over-extrapolation as observed on our jpeg evaluation design (8.9x larger than our largest training benchmark).
\begin{figure}[htbp]
    \centering
    \includegraphics[width=\columnwidth]{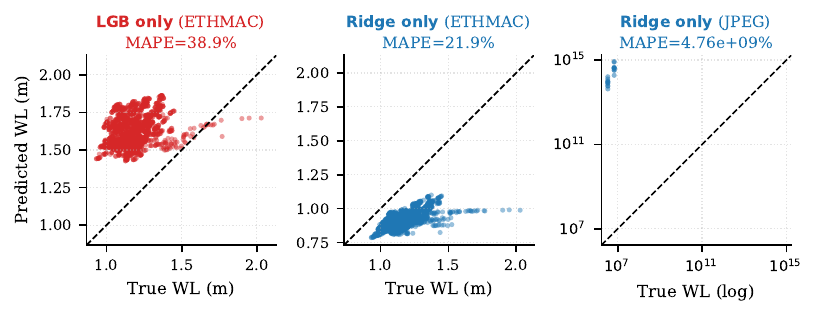}
    \caption{Predictive limitations of standalone models. (Left) LightGBM severely underpredicts the routing penalty on large training designs. (Right) Ridge Regression accurately tracks size-scaling but suffers from extreme numerical explosion when applied zero-shot to the 8.9x larger JPEG macro, justifying our blended LightGBM-Ridge architecture.}
    \label{fig:wl_scaling}
\end{figure}

Hence, we implemented a data-driven boundary safeguard based directly on the training target range. Let $[y_{\min}, y_{\max}]$ represent the absolute bounds of the log-ratio penalty observed during training. During inference, if the Ridge component predicts a raw penalty outside this known safe window, it is dynamically disabled for that specific layout. The surrogate then falls back entirely to LightGBM. Because decision trees natively cap their predictions at the maximum value seen during training, LightGBM safely bounds the routing penalty multiplier. This fallback strictly bounds the predicted penalty ($\hat{y}_{\mathrm{wl}}$) to prevent numerical explosion, while the normalizer ($\mathrm{wl}_{\mathrm{norm}}$) independently scales with the chip's massive dimensions.

\begin{table}[htbp]
\centering
\fontsize{7pt}{8pt}\selectfont 
\setlength{\tabcolsep}{3pt} 
\caption{110-Dimensional Physical Feature Space \& Nomenclature}
\label{tab:all_features_and_nom}
\vspace{1mm} 

\resizebox{\columnwidth}{!}{%
\renewcommand{\arraystretch}{1.1} 
\begin{tabular}{|c|c|c|}
\hline
\textbf{Target Head} & \multicolumn{2}{c|}{\textbf{Mathematical Formulation}} \\ \hline

\multirow{10}{*}{\textbf{\shortstack{Power\\(20)}}}
 & $\log(1 + n_{act} \times \alpha_{rel} \times f)$ & $\log(1 + n_{ff})$ \\ \cline{2-3}
 & $\alpha_{mean} / \alpha_{max}$ & $\log(1 + f_{mux} \times n_{act})$ \\ \cline{2-3}
 & $\log(1 + n_{act} \times \mu_{ds})$ & $n_{mux} / n_{act}$ \\ \cline{2-3}
 & $frac(\alpha > 2\mu_{\alpha})$ & $n_{and\_or} / n_{act}$ \\ \cline{2-3}
 & $\sigma(slack)$ & $frac(slack < 0.5\text{ns})$ \\ \cline{2-3}
 & $\mu(P_{sig=1})$ & $slack_{p10}$ \\ \cline{2-3}
 & $slack_{p50}$ & $\mu_{slack} \times f_{xor}$ \\ \cline{2-3}
 & $1 / t_{clk}$ & $Synth_{AreaWeight}$ \\ \cline{2-3}
 & $\log(1 + f_{xor} \times n_{act})$ & $\log(1 + n_{nets})$ \\ \cline{2-3}
 & $n_{nand\_nor} / n_{act}$ & $\log(1 + cd \times n_{ff} / Area)$ \\ \hline

\multirow{7}{*}{\textbf{\shortstack{Skew\\(15)}}}
 & $\log(1 + \{cd, cs, mw, bd\})$ & $cd / (spacing + 1)$ \\ \cline{2-3} 
 & $HPWL_{crit} / (cs + 1)$ & $\log(1 + HPWL_{ff})$ \\ \cline{2-3}
 & $Asymm_{crit} \times mw$ & $\rho_{crit} \times cs$ \\ \cline{2-3}
 & $die\_w / die\_h$ & $cx_{offset} \times cd$ \\ \cline{2-3}
 & $Eccentricity_{crit}$ & $\log(1 + dist_{max} / (cd+1))$ \\ \cline{2-3}
 & $\rho_{crit} / \rho_{global}$ & $\mu_y / die\_h$ \\ \hline

\multirow{42}{*}{\textbf{\shortstack{Wirelength\\(75)}}}
 & \multicolumn{2}{c|}{\textit{Layout Geometry (10)}} \\ \cline{2-3}
 & $\log(1 + n_{ff})$ & $\log(1 + Area)$ \\ \cline{2-3}
 & $\log(1 + HPWL_{ff})$ & $\log(1 + \sqrt{Area / n_{ff}})$ \\ \cline{2-3}
 & $die\_w / die\_h$ & $Aspect Ratio$ \\ \cline{2-3}
 & $cx_{ff} / die\_w$ & $cy_{ff} / die\_h$ \\ \cline{2-3}
 & $\sigma_x / die\_w$ & $\sigma_y / die\_h$ \\ \cline{2-3}
 
 & \multicolumn{2}{c|}{\textit{Cell Composition (11)}} \\ \cline{2-3}
 & $n_{xor} / n_{act}$ & $n_{mux} / n_{act}$ \\ \cline{2-3}
 & $n_{and\_or} / n_{act}$ & $n_{nand\_nor} / n_{act}$ \\ \cline{2-3}
 & $n_{ff} / n_{act}$ & $n_{buf\_inv} / n_{act}$ \\ \cline{2-3}
 & $n_{comb} / n_{ff}$ & $\mu(Drive Strength)$ \\ \cline{2-3}
 & $\sigma(Drive Strength)$ & $DS_{p90}$ \\ \cline{2-3}
 & $frac(DS \ge 4)$ & -- \\ \cline{2-3}
 
 & \multicolumn{2}{c|}{\textit{Switching Activity (8)}} \\ \cline{2-3}
 & $\log(1 + n_{act} \times \mu_{ds})$ & $\alpha_{mean} / \alpha_{max}$ \\ \cline{2-3}
 & $\mu(P_{sig=1})$ & $\sigma(\alpha) / (\mu_{\alpha}+1)$ \\ \cline{2-3}
 & $frac(\alpha = 0)$ & $frac(\alpha > 2\mu_{\alpha})$ \\ \cline{2-3}
 & $\log(1 + n_{nets})$ & $n_{nets} / n_{ff}$ \\ \cline{2-3}

 & \multicolumn{2}{c|}{\textit{Timing \& CTS Knobs (12)}} \\ \cline{2-3}
 & $1 / t_{clk}$ & $t_{clk}$ \\ \cline{2-3}
 & $Util \%$ & $\rho_{place}$ \\ \cline{2-3}
 & $cd, cs, mw, bd$ & $\log(1 + \{cd, cs, mw, bd\})$ \\ \cline{2-3} 

 & \multicolumn{2}{c|}{\textit{Knob $\times$ Circuit Interactions (12)}} \\ \cline{2-3}
 & $f_{xor} \times (n_{comb}/n_{ff})$ & $\alpha_{rel} \times f_{xor}$ \\ \cline{2-3}
 & $\alpha_{rel} \times f_{comb}$ & $\log(1 + cd \times \rho_{ff})$ \\ \cline{2-3}
 & $\log(1 + cs \times spacing)$ & $\log(1 + mw \times HPWL)$ \\ \cline{2-3}
 & $\log(1 + n_{ff} / cs)$ & $Util \times \rho_{place}$ \\ \cline{2-3}
 & $\log(1 + n_{act} \cdot \alpha \cdot f)$ & $\log(1 + n_{xor}^{act})$ \\ \cline{2-3}
 & $\log(1 + n_{mux}^{act})$ & $\log(1 + c_{ff} \times n_{ff})$ \\ \cline{2-3}

 & \multicolumn{2}{c|}{\textit{Gravity / Topology Logic-Pull ($\vec{g}$) (12)}} \\ \cline{2-3}
 & $\mu(\|\vec{g}\|)$ & $\sigma(\|\vec{g}\|)$ \\ \cline{2-3}
 & $\|\vec{g}\|_{p75}$ & $\|\vec{g}\|_{p90}$ \\ \cline{2-3}
 & $CV(\|\vec{g}\|)$ & $Gini(\|\vec{g}\|)$ \\ \cline{2-3}
 & $\mu(\|\vec{g}\| / \sqrt{Area})$ & $CV(\|\vec{g}\| / \sqrt{Area})$ \\ \cline{2-3}
 & $\|\mu_x - \mu_y\| / (\mu_x + \mu_y)$ & $\mu(\|\vec{g}\|) \times cd$ \\ \cline{2-3}
 & $\mu(\|\vec{g}\|) \times mw$ & $\mu(\|\vec{g}\|) / (spacing+1)$ \\ \cline{2-3}

 & \multicolumn{2}{c|}{\textit{Timing Path Degree \& Scale (10)}} \\ \cline{2-3}
 & $\mu(Deg_{tp})$ & $CV(Deg_{tp})$ \\ \cline{2-3}
 & $Gini(Deg_{tp})$ & $Deg_{p90}$ \\ \cline{2-3}
 & $n_{ff}^{tp} / n_{ff}^{total}$ & $n_{paths} / n_{ff}$ \\ \cline{2-3}
 & $frac(Deg > 2\mu_{deg})$ & $\log(1 + Area / n_{ff})$ \\ \cline{2-3}
 & $\log(1 + n_{comb})$ & $(n_{comb}/n_{ff}) \log(1+n_{ff})$ \\ \hline
\end{tabular}%
}

\vspace{1.5mm} 
\resizebox{\columnwidth}{!}{%
\renewcommand{\arraystretch}{1.15} 
\begin{tabular}{|c|l|c|l|}
\hline
\textbf{Sym.} & \textbf{Physical Definition} & \textbf{Sym.} & \textbf{Physical Definition} \\ \hline
$n_{act}$ & Active cell count & $cd$ & CTS Cluster Diameter \\ \hline
$n_{ff}$ & Flip-flop (sink) count & $cs$ & CTS Cluster Size \\ \hline
$n_{comb}$ & Comb. cell count & $mw$ & CTS Max Wirelength \\ \hline
$f_{xor}$ & Fraction of XOR cells & $bd$ & CTS Buffer Distance \\ \hline
$\alpha$ & Switching activity & $HPWL$ & Half-Perimeter WL \\ \hline
$P_{sig=1}$ & Logic-1 probability & $\rho_{ff}$ & FF spatial density \\ \hline
$DS$ & Cell Drive Strength & $Util$ & Core placement util. \\ \hline
$t_{clk}$ & Target clock period & $\vec{g}$ & Logic-pull (Gravity) \\ \hline
$cx, cy$ & Spatial centroids & $Deg_{tp}$ & Timing path degree \\ \hline
$\mu_x, \sigma_x$ & Mean / Std of metric $x$ & $CV$ & Coeff. of Variation \\ \hline
\end{tabular}%
}
\end{table}

 \subsubsection{Skew Prediction Head}
Clock skew is driven by extreme outliers rather than average behavior, representing the worst-case difference in clock arrival times between any two flip-flops across thousands of potential paths. Hence, unlike wirelength and power, there is
no equivalent cross-design formula that bounds the absolute skew value as it depends too heavily on the specific spatial arrangement of the flip-flops and the localized micro-decisions made by the CTS algorithm. As a consequence, achieving accurate absolute skew prediction requires grounding the model in at least one reference point. This physical anchor must be derived either from historical runs of the same placement, a similar design,  or a small number of initial calibration runs (K-shot) on the new design. By securing this statistical baseline, we can reformulate the learning objective and ask a fundamentally different question: given the spatial geometry of this specific placement's critical paths, does a given CTS configuration make the timing better or worse? To answer this, our surrogate shifts from predicting absolute nanoseconds to predicting a per-placement z-score. The target is formulated as:
\begin{equation}
    y_{z} = \frac{y_{\mathrm{ns}} - \mu}{\sigma}
\end{equation}
where $\mu$ is the mean skew for a specific placement across historical runs, and $\sigma = \max(\mathrm{std}(y_{\mathrm{ns}}), \max(|\mu| \times 0.01, 10^{-4}))$. The variance floor prevents numerical instability for placements with nearly uniform skew responses. We utilize a pure LightGBM as our learning architecture. Specifically, we extract the precise spatial coordinates of the launch and capture flip-flops for the 50 paths exhibiting the tightest timing slack prior to routing. From these coordinates, we compute dimensionless geometric ratios and explicitly interact them with the CTS routing parameters to construct our feature space (Table \ref{tab:all_features_and_nom}). This is then fed directly into the LightGBM  to predict the target z-score.

\subsection{K-Shot Calibration}
While zero-shot predictions on unseen designs achieve strong baseline accuracy, the full potential of this framework is realized when calibrated with a minimal budget of K reference runs , typically just K=1. Because the power and wirelength prediction heads are strictly tasked with regressing scale-free log-ratios, the surrogate successfully isolates and learns the universal physical sensitivities. However, it naturally suffers from a constant scalar offset caused by the absolute size and unique switching characteristics of the unseen design. 
To correct this baseline offset, we implement a $K$-shot multiplicative calibration mechanism. Given a small number of actual CTS runs ($K$) for a new placement, the framework computes a global scaling factor, $k_{\mathrm{cal}}$. To prevent the positive bias inherent in arithmetic averages, we calculate this scale using the geometric mean of the ratios via a log-space transformation, ensuring symmetric multiplicative errors cancel out:
\begin{equation}
    k_{\mathrm{cal}} = \exp \left( \frac{1}{K} \sum_{i=1}^{K} \log \left( \frac{y_{\mathrm{true}}^{(i)}}{\hat{y}_{\mathrm{pred}}^{(i)}} \right) \right)
\end{equation}

This scalar is then broadcast across all subsequent predictions for that specific placement for accurate prediction:
\begin{equation}
    \hat{y}_{\mathrm{calibrated}} = k_{\mathrm{cal}} \times \hat{y}_{\mathrm{pred}}
\end{equation}

 \subsection{Pareto Optimization}

  We explore the design space by tweaking four CTS knobs within the physical bounds of our training data. Because any single CTS configuration simultaneously determines power, wirelength, skew, no single configuration can minimize all three objectives simultaneously. Thus, we seek to construct the Pareto front. 
 However, finding the optimal configuration is a massive search and compute problem involving millions of potential knob combinations. We solve this by a key structural property of our surrogate: for a placement, every feature derived from the layout geometry, switching activity, and timing reports is a strict constant. By isolating the four CTS knobs and their derived features as the only dynamic variables during exploration, we avoid the overhead of full feature extraction. Instead, we construct the batch feature matrix by simply tiling a single base layout vector $N$ times and patching the knob-dependent columns using vectorized operation. This dynamically patched matrix is then fed directly into the native inference engines of the ensembles of the prediction head. 
 To extract the final front, the predicted values are assembled into a cost matrix $\mathbf{C} \in \mathbb{R}^{N \times 3}$ and normalized to a $[0, 1]$ range. A candidate $i$ is identified as dominated by candidate $j$ if it is worse across all metrics:
$$\forall k \in \{1, \dots, 3\}, \; \mathbf{C}_{j,k} \leq \mathbf{C}_{i,k} + \varepsilon \quad \text{and} \quad \exists k : \mathbf{C}_{j,k} < \mathbf{C}_{i,k} - \varepsilon$$
Crucially, our framework is algorithm-agnostic. The architecture is designed to seamlessly plug in any advanced search algorithm such as NSGA-II to intelligently navigate toward the most promising regions and find optimal configurations.


\section{Experiments}
All experiments were conducted on a machine with an Intel Core i7-14700HX CPU (14 cores, 28 threads), 16 GB RAM.

\subsection{Dataset Generation and Benchmarks}
We generated a comprehensive dataset of 5,520 unique CTS evaluations derived from six distinct open-source IP cores of varying scale, logic depth, and functional complexity utilizing CTS-Bench \cite{khadka2026}  with Sky130 PDK and OpenROAD \cite{OpenRoad}. To rigorously evaluate cross-design generalization, the corpus is strictly partitioned into a training set and an OOD test set.  The training dataset comprises 5,400 CTS runs across four base architectures (AES, PicoRV32, SHA-256, and ETHmac), originating from 540 unique pre-CTS physical placements. The OOD test set is held completely unseen during training and consists of 120 additional CTS runs across two structurally distinct architectures (60 runs each for JPEG Encoder and ZipDiv Core, spanning 6 unique placements each). To effectively capture the localized Pareto surface for every design, each individual placement in both the training and test sets was routed under 10 distinct CTS configurations. These configurations were generated by uniformly sampling a design space of four primary CTS tuning knobs. The constraints and parameters varied to generate this dataset, along with their sampled ranges, are detailed in Table \ref{tab:params}.

\begin{table}[h]
\caption{Randomization Knobs for Data Diversity}
\label{tab:params}
\centering
\small
\begin{tabular}{lll}
\toprule
\textbf{Stage} & \textbf{Parameter} & \textbf{Range / Set} \\
\midrule
Synthesis & Synth Strategy & \{AREA 0-2, DELAY 0-4\} \\
Floorplan & Aspect Ratio & \{0.7, 1.0, 1.4, 2.0\} \\
Floorplan & IO Mode & \{Random Pin, Standard\} \\
Placement & Core Utilization & 40\% -- 70\% \\
Placement & Target Density & $Util + [0.0, 0.20]$ \\
Placement & Time Driven & \{Enabled, Disabled\} \\
Placement & Routability Driven & \{Enabled, Disabled\} \\
\midrule
CTS & Sink Max Dia. & 35 -- 70 $\mu m$ \\
CTS & Max Wire Length & 130 -- 280 $\mu m$ \\
CTS & Cluster Size & 12 -- 30 sinks \\
CTS & Buffer Distance & 70 -- 150 $\mu m$ \\
\bottomrule
\end{tabular}
\end{table}

\begin{table*}[tbp]
\centering
\small
\renewcommand{\arraystretch}{1.15}
\setlength{\tabcolsep}{5pt}
\caption{SwiftCTS Outcomes Prediction Results Across All Benchmarks. Validated on Unseen Designs (LODO) and OOD Macros.}
\label{tab:master_results}
\begin{tabular}{|c|c|rrrr|rrrr|rrrr|}
\hline
\multirow{2}{*}{\begin{tabular}{@{}c@{}}\textbf{Evaluation} \\ \textbf{Mode}\end{tabular}} & 
\multirow{2}{*}{\begin{tabular}{@{}c@{}}\textbf{Unseen} \\ \textbf{Placement}\end{tabular}} & 
\multicolumn{4}{c|}{\textbf{Power (MAPE \%)}} & 
\multicolumn{4}{c|}{\textbf{Wirelength (MAPE \%)}} & 
\multicolumn{4}{c|}{\textbf{Achieved Skew (MAE ns)}} \\
 & & K=0 & K=1 & K=2 & K=5 & K=0 & K=1 & K=2 & K=5 & K=0 & K=1 & K=2 & K=5 \\ \hline
\multirow{5}{*}{\shortstack{\textbf{LODO}\\(Base)}} 
 & AES & 21.0 & 2.2 & 2.0 & \textbf{1.8} & 17.6 & 0.7 & 0.6 & \textbf{0.5} & -- & 0.1525 & 0.1174 & \textbf{0.0927} \\
 & ETHMAC & 5.7 & 5.0 & 4.3 & \textbf{4.0} & 8.0 & 0.8 & 0.7 & \textbf{0.7} & -- & 0.1198 & 0.0986 & \textbf{0.0834} \\
 & \cellcolor{bestrow}PicoRV32 & \cellcolor{bestrow}16.8 & \cellcolor{bestrow}2.3 & \cellcolor{bestrow}2.1 & \cellcolor{bestrow}\textbf{1.8} & \cellcolor{bestrow}8.0 & \cellcolor{bestrow}0.4 & \cellcolor{bestrow}0.4 & \cellcolor{bestrow}\textbf{0.3} & \cellcolor{bestrow}-- & \cellcolor{bestrow}0.0953 & \cellcolor{bestrow}0.0802 & \cellcolor{bestrow}\textbf{0.0665} \\
 & SHA256 & 54.4 & 3.9 & 3.4 & \textbf{3.2} & 7.5 & 0.4 & \textbf{0.3} & \textbf{0.3} & -- & 0.0982 & 0.0853 & \textbf{0.0713} \\ \cline{2-14}
 & \textbf{Mean} & 24.5 & 3.3 & 2.9 & \textbf{2.7} & 10.3 & 0.6 & 0.5 & \textbf{0.5} & -- & 0.1164 & 0.0954 & \textbf{0.0785} \\ \hline
\multirow{7}{*}{\shortstack{\textbf{OOD}\\(Large)}} 
 & JPEG-1 & 23.9 & \textbf{3.6} & 4.1 & 4.5 & 22.0 & 2.0 & 1.0 & \textbf{0.7} & -- & 0.2072 & 0.2604 & \textbf{0.2020} \\
 & JPEG-2 & 44.1 & \textbf{6.5} & 7.9 & 7.3 & 7.8 & \textbf{1.1} & 1.2 & \textbf{1.1} & -- & \textbf{0.1366} & 0.1595 & 0.1385 \\
 & \cellcolor{bestrow}JPEG-3 & \cellcolor{bestrow}19.9 & \cellcolor{bestrow}3.6 & \cellcolor{bestrow}\textbf{3.4} & \cellcolor{bestrow}4.7 & \cellcolor{bestrow}0.8 & \cellcolor{bestrow}1.0 & \cellcolor{bestrow}\textbf{0.8} & \cellcolor{bestrow}\textbf{0.8} & \cellcolor{bestrow}-- & \cellcolor{bestrow}0.1149 & \cellcolor{bestrow}0.1208 & \cellcolor{bestrow}\textbf{0.1020} \\
 & JPEG-4 & 37.6 & 3.3 & \textbf{1.6} & 1.7 & 8.3 & \textbf{1.1} & 1.2 & \textbf{1.1} & -- & 0.5434 & 0.2692 & \textbf{0.1849} \\
 & JPEG-5 & 41.8 & 5.3 & \textbf{2.8} & 4.5 & 1.0 & 1.8 & \textbf{0.5} & 0.6 & -- & 0.2648 & 0.0946 & \textbf{0.0937} \\
 & JPEG-6 & 19.9 & \textbf{4.0} & 4.7 & \textbf{4.0} & 16.8 & 1.6 & 2.0 & \textbf{1.1} & -- & 0.2258 & \textbf{0.1652} & 0.1995 \\ \cline{2-14}
 & \textbf{Mean} & 31.2 & 4.4 & \textbf{4.1} & 4.5 & 9.4 & 1.4 & 1.1 & \textbf{0.9} & -- & 0.2488 & 0.1783 & \textbf{0.1534} \\ \hline
\multirow{7}{*}{\shortstack{\textbf{OOD}\\(Small)}} 
 & Zipdiv-1 & 7.2 & \textbf{5.5} & 6.5 & 7.9 & 63.6 & 0.8 & 0.9 & \textbf{1.1} & -- & \textbf{0.0015} & 0.0016 & 0.0019 \\
 & Zipdiv-2 & 10.9 & 6.4 & \textbf{4.5} & 5.5 & 35.6 & 0.7 & 0.7 & \textbf{0.6} & -- & 0.0038 & 0.0030 & \textbf{0.0018} \\
 & Zipdiv-3 & 9.7 & 2.3 & 3.2 & \textbf{0.5} & 46.3 & 0.6 & 0.3 & \textbf{0.1} & -- & 0.0048 & \textbf{0.0035} & 0.0042 \\
 & \cellcolor{bestrow}Zipdiv-4 & \cellcolor{bestrow}4.6 & \cellcolor{bestrow}\textbf{2.3} & \cellcolor{bestrow}2.8 & \cellcolor{bestrow}2.8 & \cellcolor{bestrow}60.4 & \cellcolor{bestrow}\textbf{0.4} & \cellcolor{bestrow}0.6 & \cellcolor{bestrow}\textbf{0.4} & \cellcolor{bestrow}-- & \cellcolor{bestrow}0.0024 & \cellcolor{bestrow}0.0023 & \cellcolor{bestrow}\textbf{0.0018} \\
 & Zipdiv-5 & 10.6 & 6.8 & 9.7 & \textbf{4.3} & 70.0 & 1.2 & 1.3 & \textbf{0.7} & -- & 0.0066 & 0.0037 & \textbf{0.0026} \\
 & Zipdiv-6 & 7.4 & \textbf{4.7} & 8.0 & 6.5 & 63.6 & \textbf{0.5} & 0.7 & 0.6 & -- & \textbf{0.0040} & 0.0043 & 0.0055 \\ \cline{2-14}
 & \textbf{Mean} & 8.4 & 4.7 & 5.8 & \textbf{4.6} & 56.6 & 0.7 & 0.7 & \textbf{0.6} & -- & 0.0038 & 0.0031 & \textbf{0.0030} \\ \hline
\end{tabular}
\vspace{1.0mm}
\begin{minipage}{\linewidth}
\small $^\dagger$Skew MAE is measured in nanoseconds (ns). K=0 yields relative z-scores; absolute ns extraction requires K$\ge$1 to anchor the distribution.
\end{minipage}
\end{table*}

\subsection{Two-Stage Evaluation Strategy}
To rigorously evaluate the framework's ability to generalize, we employ a strict two-stage protocol. 

\paragraph{Stage 1: Leave-One-Design-Out (LODO) Cross-Validation.} We first conduct an internal out-of-distribution test using the four base architectures. In each training fold, the surrogate is trained on placements from only three design families and tested entirely zero-shot on the held-out design. This stage proves that SwiftCTS can learn highly accurate predictive models from an extremely limited data volume (just three designs) without overfitting to their specific logic paths.

\paragraph{Stage 2: Out-of-Distribution (OOD)} While LODO demonstrates baseline generalization, it could be argued that the overall training process or feature selection might remain implicitly biased toward the structural characteristics of the four base designs. To definitively prove robustness, we train a final surrogate on the complete corpus of the four base architectures and deploy it zero-shot on two completely distinct OOD IPs. We deliberately selected ZipDiv Core (21× smaller than smallest training design) and JPEG Encode (5× larger than the largest training design) to test the framework to its absolute limits.

\subsection{End-to-End Pareto Optimization and Search Efficiency}
To demonstrate the practical utility of our ultra-fast inference engine for DSE, we evaluate its ability to navigate a massive combinatorial search space. Discretizing our four primary CTS tuning knobs yields a grid of approximately 8.36 million unique configurations. We wrap the SwiftCTS prediction heads within an NSGA-II evolutionary optimizer, constraining the search to a budget of 100K evaluations. To benchmark both search efficiency and the optimality of the resulting Pareto frontier, we compare the NSGA-II framework against baseline uniform Random Search and Sobol sequence sampling. Because SwiftCTS maintains exceptionally low underlying prediction errors, dynamically tiling the static layout context with varying CTS knobs produces a highly reliable cost matrix for the optimizer. Finally, to validate true QoR and confirm that the surrogate's mathematical predictions align with physical reality, the optimal configurations identified by the search engine are fed back into the OpenROAD \cite{OpenRoad} physical design flow. We then directly compare the surrogate's predicted metrics against the actual routed outcomes.


\section{Results}

To evaluate SwiftCTS, we report the Mean Absolute Percentage Error (MAPE) for clock power and wirelength, and the Mean Absolute Error (MAE) in nanoseconds for timing skew. We use MAPE for power and wirelength to normalize the massive scale differences between varying macro architectures. Conversely, we use MAE for skew to provide a strict, absolute measure of worst-case temporal error. All predictions are subsequently validated against actual OpenROAD runs.
\subsection{Predictions on LODO and OOD}
The practical utility of the SwiftCTS framework depends directly on the fidelity of its surrogate predictions.  Hence, to validate this predictive baseline, Table \ref{tab:master_results} reports prediction fidelity across both evaluation stages. Under zero-shot conditions ($K=0$), the surrogate successfully captures the relative sensitivities of the CTS knobs but predictably suffers from absolute size and density offsets. This is particularly evident in the exceptionally dense ZipDiv macro, which exhibits a 56.6\% zero-shot wirelength error due to severe domain shift from the training distribution. However, applying a single calibration run ($K=1$) mathematically anchors the predictions and virtually eliminates this shift. Across the LODO base designs, $K=1$ calibration drops power prediction error to 3.3\% and wirelength error to an exceptional 0.6\%. Most notably, this fidelity is preserved on the extreme OOD benchmarks. On the massive JPEG encode macro, wirelength prediction error drops to 1.4\% with just one physical reference. Furthermore, while skew prediction fundamentally requires at least one physical anchor ($K\ge1$) to convert relative spatial z-scores into absolute temporal values, the model consistently bounds timing prediction errors to approximately 0.11 ns for base designs, and mere picoseconds for the smaller ZipDiv core. This confirms that SwiftCTS yields near-physical accuracy across highly diverse topologies, making it a reliable engine for driving aggressive combinatorial search.

\subsection{Search Dynamics and Algorithmic Efficiency}
Although SwiftCTS accelerates individual CTS predictions to sub-millisecond latencies, the overall runtime of Design Space Exploration remains bounded by the chosen search algorithm. Baseline heuristics, such as Random Search or Sobol sequences, sample the design space uniformly. and lack a convergence criterion. In contrast, evolutionary algorithms like NSGA-II trades the guarantee of exhaustive search for accelerated convergence. As illustrated in Table \ref{tab:search_time}, SwiftCTS enables NSGA-II to evaluate a massive budget of 100,000 configurations in an average of only 8.9 seconds. However, one inherent risk of evolutionary algorithms like NSGA-II is their dependence on the initial starting population. If the initial random guesses are poor or clustered in a sub-optimal region, the algorithm can prematurely converge on a localized, inferior solution rather than the true global optimum. However, the instantaneous inference speed of SwiftCTS provides a direct solution to this vulnerability. Because a full 100,000-evaluation search takes under 10 seconds, engineers can effortlessly launch multiple differently-seeded searches in parallel. This multi-start approach guarantees a highly robust global Pareto front while remaining orders of magnitude faster than a single conventional Random Search. 
\begin{table}[htbp]
\centering
\caption{SwiftCTS DSE Runtime Comparison (100,000 Evaluations)}
\label{tab:search_time}
\resizebox{\columnwidth}{!}{%
\renewcommand{\arraystretch}{1.25} 
\begin{tabular}{@{}lll@{}}
\toprule
\textbf{Search Algorithm} & \textbf{Convergence Profile} & \textbf{Runtime (Avg.)} \\ \midrule
Random Search & Exhaustive Coverage & 604.3s \\
Sobol Sequence & Exhaustive Coverage & 600.0s \\
\textbf{NSGA-II (Single)} & \textbf{Evolutionary} & \textbf{8.9s}\\ \bottomrule
\end{tabular}%
}
\end{table}

\begin{figure}[htbp] 
    \centering
    \includegraphics[width=\columnwidth]{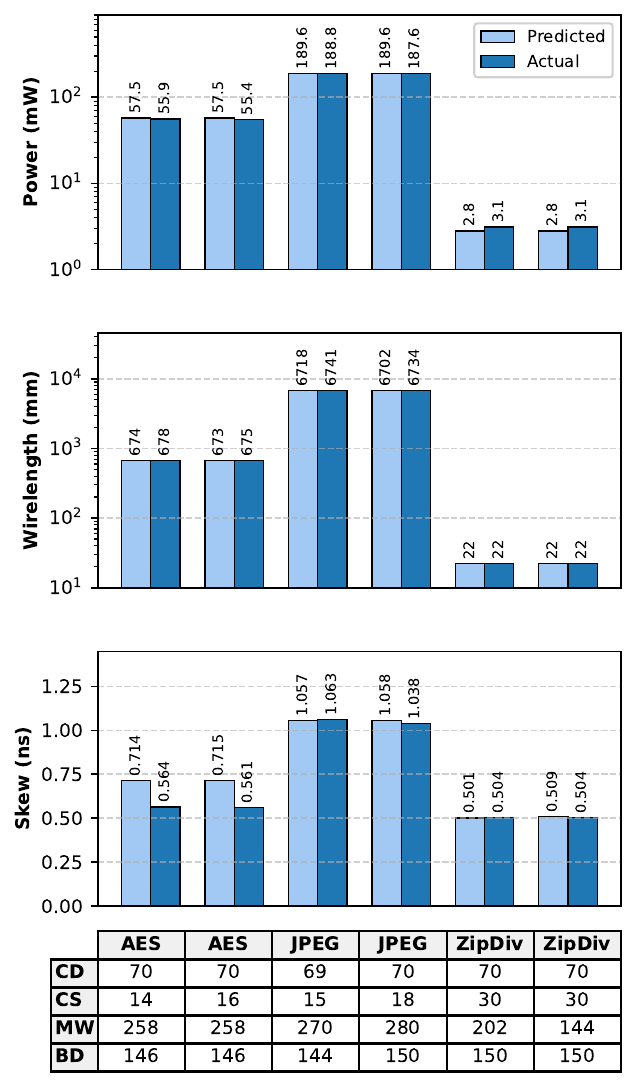} 
    \caption{Closed-loop physical validation comparing SwiftCTS predictions against actual OpenROAD routing. The integrated data table details the Pareto-optimal CTS configurations evaluated for each top-ranked design.}
    \vspace{-20pt}
    \label{fig:physical_validation}
\end{figure}
\begin{table*}[t]
\centering
\caption{Feature and Performance Comparison of ML-Driven CTS Frameworks}
\label{tab:comparison}
\renewcommand{\arraystretch}{1.3} 
\setlength{\tabcolsep}{5pt} 
\resizebox{\textwidth}{!}{
\begin{tabular}{@{}llcllccc@{}}
\toprule
\multirow{2}{*}{\textbf{Framework (Arch.)}} & \multirow{2}{*}{\textbf{Input Data}} & \multirow{2}{*}{\textbf{GPU?}} & \multirow{2}{*}{\textbf{Train Time}} & \multirow{2}{*}{\textbf{DSE Scale}} & \multicolumn{3}{c}{\textbf{Prediction Error}} \\ \cmidrule(l){6-8} 
 &  &  &  &  & \textbf{Power} & \textbf{WL} & \textbf{Skew} \\ \midrule
Kwon et al. \cite{kwon2018} (ANN) & Components & Yes & N/R & None (Pred. Only) & N/R & 13--17\%$^\ast$ & N/R \\
Nagaria et al. \cite{nagaria2020} (CNN) & Spatial Images & Yes & N/R & 1-Shot LP & 14.89\% & 8--15\%$^\ast$ & N/R \\
GAN-CTS \cite{gan-cts} (GAN) & Spatial Images & Yes & $\sim$6 hrs & Probabilistic 1-Shot & 1.26\% & 1.64\% & 3.57\% \\ \midrule
\textbf{SwiftCTS (GBDT)} & \textbf{Physical Stats} & \textbf{No} & \textbf{4.1 sec} & \textbf{Comprehensive (100k in 8.9s)} & \textbf{3.3\%} & \textbf{0.6\%} & \textbf{0.11 ns} \\ \bottomrule
\end{tabular}%
}
\begin{minipage}{\textwidth}
\vspace{2mm}
\raggedright \footnotesize
N/R: Not Reported. $^\ast$Error represents intermediate component estimation (e.g., buffer count, wireload) rather than final routed metric.
\end{minipage}
\vspace{-2mm}
\end{table*}
\subsection{Closed-Loop Physical Validation}
\begin{figure}[htbp]
    \centering
    \includegraphics[width=0.9\linewidth]{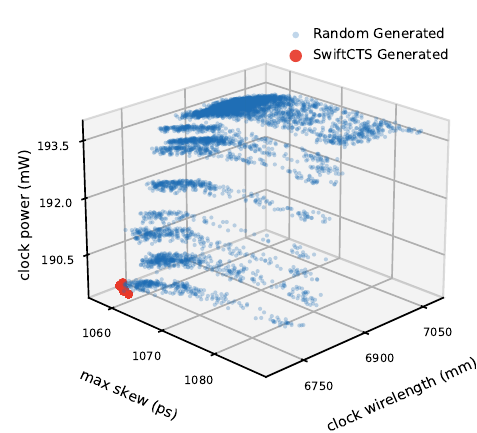}
    \vspace{-6pt}
    \caption{Design space exploration outcomes on the unseen JPEG macro.}
    \vspace{-8pt}
    \label{fig:pareto_jpeg}
\end{figure}

To verify that the optimal configurations found during the search process work in practice, we perform a closed-loop validation. From the 100,000 configurations evaluated by the search engine, we select the top-performing configurations for each design and execute them through the actual OpenROAD flow. First, we compare the surrogate model's predictions directly against the actual physical outputs from OpenROAD (Fig \ref{fig:physical_validation}). The results show strong alignment across all test cases, proving that the search algorithm is finding genuine physical improvements. For example, on the completely unseen JPEG Encode design, the actual routed power (188.80 mW) matches the prediction (189.63 mW) with less than 0.5\% error. Similarly, for the ZipDiv core, the predicted clock skew is within 5 picoseconds of the actual physical routing. Second, we compare these validated SwiftCTS results against the default OpenROAD settings. Across all benchmarks, the configurations found by our framework and the resulting clock tree constructed is superior to baseline run (Fig \ref{fig:layout_comp}). These results confirm that the framework provides highly accurate predictions and guides it towards superior layouts. (Fig \ref{fig:pareto_jpeg})

\section{Discussion}

\subsection{Comparison to existing frameworks}

\begin{figure}[htbp]
    \centering
    \includegraphics[width=\columnwidth]{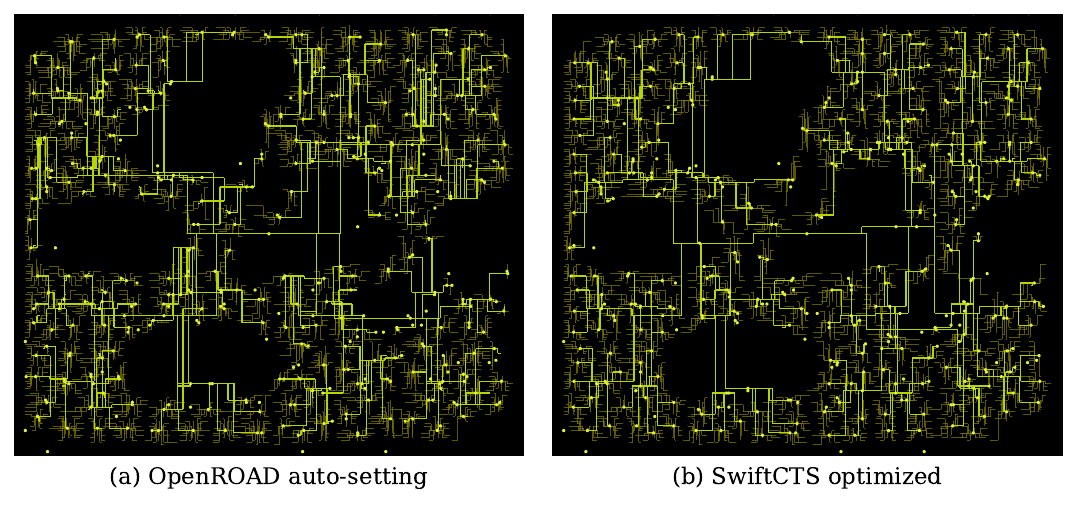}
    \caption{Clock tree layout comparison on the AES benchmark, highlighting the relaxation of congestion after applying the suggested SwiftCTS parameters.}
    \label{fig:layout_comp}
\end{figure}

As summarized in Table \ref{tab:comparison}, SwiftCTS addresses the fundamental reliability and efficiency gaps found in current frameworks. While GAN-CTS demonstrates impressive power and wirelength reductions, its reliance on a probabilistic, one-shot suggestion is the biggest hurdle towards reliable Design Space Exploration (DSE). For instance, GAN-CTS reported a nearly 45\% failure rate on their NOVA benchmark, alongside numerous False Successes where the model recommended configurations worse than default tools \cite{gan-cts}. SwiftCTS eliminates these vulnerabilities simply by invoking the prediction engine 100,000 times for every placement and actively evaluating thousands of different knob combinations on the current layout with the prediction engine and ensures that there is no possibility of failed runs or False Successes, provided that the underlying prediction is accurate. Consequently, SwiftCTS consistently finds configurations that perform strictly better than baseline heuristics across all target metrics, as validated by the physical outcomes in Fig. \ref{fig:layout_comp}.

\subsection{The Necessity of Placement-Level Calibration}
A theoretical alternative to our per-placement calibration methodology is an architecture-level approach: executing $K$ runs on a single physical placement and broadcasting that computed scalar ($k_{\mathrm{cal}}$) across all other unseen placements of the same design. To rigorously test this, we conducted an ablation study across the LODO cross-validation folds. For each held-out design, we randomly sampled $K$ reference placements to compute a global scale, and evaluated the surrogate on all remaining unseen placements, averaging the results over multiple random seeds. Table \ref{tab:ablation_cp} reports the MAPE and its standard deviation ($\pm$) averaged across all held-out LODO cross-validation designs. These results demonstrate the critical limitations of architecture-level calibration and points to a fundamental reality that clock tree metrics are highly governed by the physical floorplan rather than the logical netlist alone. Applying a scalar derived from a congested, center-clustered floorplan to a dispersed floorplan introduces substantial baseline errors. These findings justify the necessity of our placement-level calibration methodology. 

\begin{table}[htbp]
\centering
\caption{Ablation: Cross-Placement (Architecture-Level) Calibration}
\label{tab:ablation_cp}
\begin{tabular}{@{}lcccc@{}}
\toprule
\textbf{MAPE} & $\mathbf{K=0}$ & $\mathbf{K=1}$ & $\mathbf{K=5}$ & $\mathbf{K=10}$ \\ \midrule
\textbf{Power} & 24.5\% & 13.6\% \small{$\pm 8.2\%$} & 11.2\% \small{$\pm 6.0\%$} & 10.4\% \small{$\pm 5.3\%$} \\
\textbf{WL} & 10.3\% & 11.4\% \small{$\pm 6.4\%$} & 9.9\% \small{$\pm 4.7\%$} & 9.2\% \small{$\pm 4.5\%$} \\
\bottomrule
\multicolumn{5}{@{}p{0.45\textwidth}@{}}{\small \textit{}}
\end{tabular}
\end{table}

\section{Conclusion}
In this paper, we proposed SwiftCTS, a physics-informed surrogate framework utilizing gradient-boosted ensembles to accelerate CTS design space exploration. By employing XGBoost and LightGBM, SwiftCTS enables training from scratch in under five seconds on a standard CPU and delivers sub-millisecond per-prediction inference. Utilizing K-shot calibration, we anchor predictions to unseen designs with minimal data; a single reference run ($K=1$) reduces power prediction error from 24.5\% to 3.3\% and wirelength error from 56.6\% to under 1\%. When integrated with an evolutionary optimizer, the framework evaluates 100,000 configurations in an average of 8.9 seconds, with timing skew predictions validated within five picoseconds of actual physical routing. Despite these gains, limitations remain for industrial adoption. The current feature space does not contain defined attributes like dont-touch nets or localized hold-only optimizations. Furthermore, while validated on the OpenROAD flow and Sky130, the framework’s robustness across advanced industrial technology nodes and commercial suites like Innovus or ICC2 requires further verification. Future work will incorporate multi-Vt designs , non-uniform power grids and industrial designs to better reflect modern industrial complexity.

\bibliographystyle{IEEEtran}
\bibliography{references}

\end{document}